# Sentiment Analysis on Speaker Specific Speech Data

Maghilnan S, Rajesh Kumar M, Senior IEEE, Member
School of Electronic Engineering
VIT University
Tamil Nadu, India
maghilnan.s2013@vit.ac.in, mrajeshkumar@vit.ac.in

*Abstract*— Sentiment analysis has evolved over past few decades, most of the work in it revolved around textual sentiment analysis with text mining techniques. But audio sentiment analysis is still in a nascent stage in the research community. In this proposed research, we perform sentiment analysis on speaker discriminated speech transcripts to detect the emotions of the individual speakers involved in the conversation. We analyzed different techniques to perform speaker discrimination and sentiment analysis to find efficient algorithms to perform this task.

## I. INTRODUCTION

Sentiment Analysis is the study of people's emotion or attitude towards a event, conversation on topics or in general. Sentiment analysis is used in various applications, here we use it to comprehend the mindset of humans based on their conversations with each other. For a machine to understand the mindset/mood of the humans through a conversation, it needs to know who are interacting in the conversation and what is spoken, so we implement a speaker and speech recognition system first and perform sentiment analysis on the data extracted from prior processes.

Understanding the mood of humans can be very useful in many instances. For example, computers that possess the ability to perceive and respond to human non-lexical communication such as emotions. In such a case, after detecting humans' emotions, the machine could customize the settings according his/her needs and preferences.

The research community has worked on transforming audio materials such as songs, debates, news, political arguments, to text. And the community also worked on audio analysis investigation [1,2,3] to study customer service phone conversations and other conversations which involved more than one speaker. Since there is more than one speaker involved in the conversation it becomes clumsy to do analysis on the audio recordings, so in this paper we propose a system which would be aware of the speaker identity and perform audio analysis for individual speakers and report their emotion.

The approach followed in the paper investigates the challenges' and methods to perform audio sentiment analysis on audio recordings using speech recognition and speaker recognition. We use speech recognition tools to transcribe the audio recordings and a proposed speaker discrimination method based on certain hypothesis to identify the speakers involved in a conversation. Further, sentiment analysis is performed on the speaker specific speech data which enables the machine to understand what the humans were talking about and how they feel.

Section-II discusses the theory behind Speaker, Speech Recognition and Sentiment Analysis is discussed. Section-III contains explanation about the proposed system. Section-IV contains details about the experimental setup and Section-V presents result obtained and detailed analysis. The work is concluded in Section-VI.

## II. RELATED WORK AND BACKGROUND

*A. Sentiment Analysis:*

Sentiment Analysis, shortly referred as SA, which identifies the sentiment expressed in a text then analyses it to find whether document expresses positive or negative sentiment. Majority of work on sentiment analysis has focused on methods such as Naive Bayesian, decision tree, support vector machine, maximum entropy [1,2,3]. In the work done by Mostafa et al [4] the sentences in each document are labelled as subjective and objective (discard the objective part) and then classical machine learning techniques are applied for the subjective parts. So that the polarity classifier ignores the irrelevant or misleading terms. Since collecting and labelling the data is time consuming at the sentence level, this approach is not easy to test. To perform sentiment analysis, we have used the following methods – Naive Bayes, Linear Support Vector Machines, VADER [6]. And a comparison is made to find the efficient algorithm for our purpose.

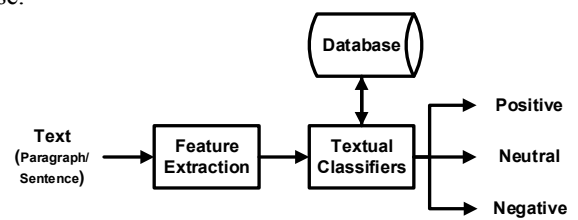

Fig. 1. Framework of Generic Sentiment Analysis System

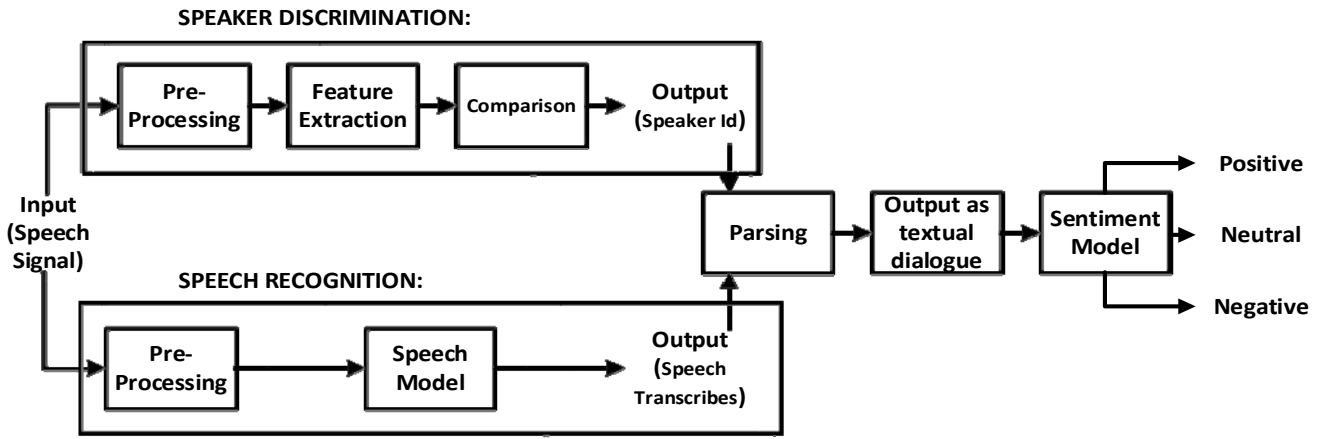

Fig. 2. Proposed Structure for the Sentiment Analysis System

*B. Speech Recognition:*

Speech recognition is the ability given to a machine or program to identify words and phrases in language spoken by humans and convert them to a machine-readable format, which can be further used for processing. In this paper, we have used speech recognition tools such as Sphinx4 [5], Bing Speech, Google Speech Recognition. A comparison is made and the best suite for the proposed model is chosen.

*C. Speaker Recognition:*

Identifying a human based on the variations and unique characteristics in the voice is referred to speaker recognition. It has acquired a lot of attention from the research community for almost eight decades [7]. Speech as signal contains several features which can extract linguistic, emotional, speaker specific information [8], speaker recognition harnesses the speaker specific features from the speech signal.

In this paper, Mel Frequency Cepstrum Coefficient (MFCC) is used for designing a speaker discriminant system. The MFCC's for speech samples from various speakers are extracted and compared with each other to find the similarities between the speech samples.

1) Feature Extraction:

The extraction of unique speaker discriminant feature is important to achieve a better accuracy rate. The accuracy of this phase is important to the next phase, because it acts as the input for the next phase.

*MFCC*— Humans perceive audio in a nonlinear scale, MFCC tries to replicate the human ear as a mathematical model. The actual acoustic frequencies are mapped to Mel frequencies which typically range between 300Hz to 5KHz. The Mel scale is linear below 1KHz and logarithmic above 1KHz. MFCC Constants signifies the energy associated with each Mel bin, which is unique to every speaker. This uniqueness enables us to identify speakers based on their voice [9].

2) Feature Matching:

***Dynamic Time Wrapping(DTW)***— Stan Salvador et al [7] describes DTW algorithm as Dynamic Programming techniques. This algorithm measures the similarity between two time series which varies in speed or time. This technique is also used to find the optimal alignment between the times series if one time series may be "warped" non-linearly by stretching or shrinking it along its time axis. This warping between two time series can then be used to find corresponding regions between the two time series or to determine the similarity between the two time series. The principle of DTW is to compare two dynamic patterns and measure its similarity by calculating a minimum distance between them.

Once the time series is wrapped, various distance/similarity computation methods such as Euclidean distance, Canberra Distance, Correlation can be used. A comparison between these methods is shown in results section.

### III. PROPOSED SYSTEM

In this paper, we propose a model for sentiment analysis that utilizes features extracted from the speech signal to detect the emotions of the speakers involved in the conversation. The process involves four steps: 1) Pre-processing which includes VAD, 2) Speech Recognition System, 3) Speaker Recognition System, 4) Sentiment Analysis System.

The input signal is passed to the Voice Activity Detection System, which identifies and segregates the voices from the signal. The voices are stored as chunks in the database, the chunks are then passed to speech recognition and speaker discrimination system for recognizing the content and speaker Id. Speaker recognition system tags the chunks with the speaker ids, it should be noted that the system works in an unsupervised fashion, i.e. it would find weather the chunks are from same speaker or different and tag it as 'Speaker 1' and 'Speaker 2'. The speech recognition system transcribes the chunks to text. The system further matches the Speaker Id with the transcribed text. It is stored as dialogue in the database. The text output from the speech recognition system specific to individual speaker serves as potential feature to estimate sentiment emphasized by the individual speaker. The entire process is depicted pictorially in Figure 2.

## IV. EXPERIMENTAL SETUP

### A. Dataset

Our dataset comprises of 21 audio files recorded in a controlled environment [10]. Three different scripts are used as conversation between two peoples. Seven speakers are totally involved in these recordings, 4 males and 3 females. The conversations are prelabelled depending upon the scenario. The audio is sampled at 16KHz and recorded as mono tracks for an average of 10 seconds.

A dataset sample is shown in Figure 3

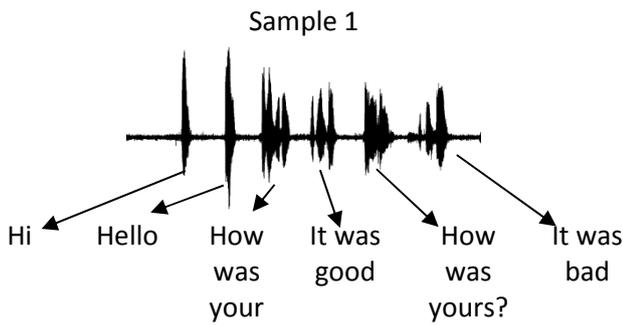

Fig. 3.  Sample Waveform

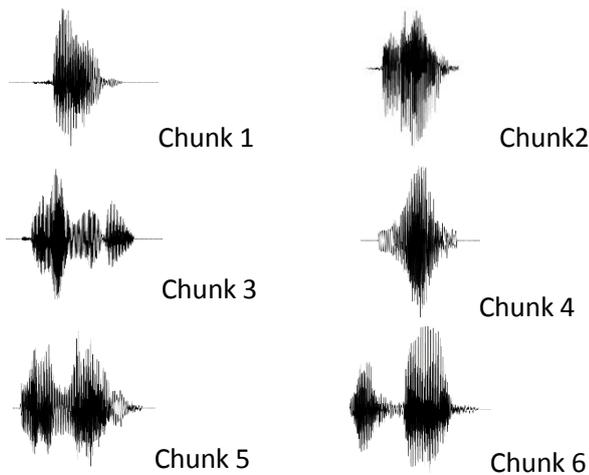

Fig. 4.  After segmentation with VAD

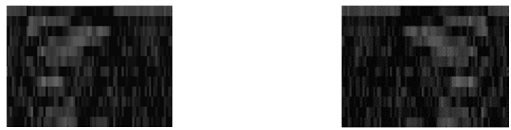

Fig. 5.  MFCC feature of Chunk1 and Chunk2

### B. Experiments and Evaluation Metrics:

The proposed system uses speech, speaker recognition and sentiment analysis. We have presented a detailed analysis for the experiments performed with various tools and algorithms. The tools used for speech recognition are Sphinx4, Bing Speech API, Google Speech API. And performance metric used was WWR. For speaker recognition, we used MFCC as feature and DTW with various distance computation methods such as Euclidean, Correlation, Canberra for feature matching. And recognition rate is used as the performance metric. For sentiment analysis, standard sentiment analysis datasets viz. twitter dataset, product review dataset [6] are used to commute the accuracy of the system.

## V. RESULTS

### A. Results for Automatic Speech Recognition Engine:

First, the audio files from the dataset are converted to text files through different speech recognition tools. Table 1 shows the WRR obtained for various scripts which were spoken by different speakers. M1 refers to Male speaker 1, similarly F1 refers to Female speaker 1. The WRR is given as percentage values.

TABLE I. WRR for Sphinx4:

| Speaker1 | Speaker2 | Script1 | Script2 | Script3 |
|---|---|---|---|---|
| M1 | M2 | 46.67 | 23.08 | 62.50 |
| M2 | M3 | 33.33 | 30.77 | 18.75 |
| M3 | M1 | 26.67 | 23.08 | 25.00 |
| M2 | M4 | 53.33 | 30.77 | 56.25 |
| F1 | F2 | 46.67 | 38.46 | 31.25 |
| F2 | F3 | 26.67 | 30.77 | 37.50 |
| F3 | F1 | 33.33 | 38.46 | 25.00 |

Table 2, tabulates the WRR obtained by using Google Speech API to transcribe the speech signals. The same dataset is used i.e. the same scripts and the same persons are used to compare the results. This is done to validate to tools in equal basis.

TABLE II. WRR for Google Speech API:

| Speaker 1 | Speaker 2 | Script1 | Script2 | Script3 |
|---|---|---|---|---|
| M1 | M2 | 93.33 | 84.62 | 81.25 |
| M2 | M3 | 86.67 | 92.31 | 75.00 |
| M3 | M1 | 86.67 | 84.62 | 43.75 |
| M2 | M4 | 80.00 | 76.92 | 68.75 |
| F1 | F2 | 86.67 | 84.62 | 81.25 |
| F2 | F3 | 93.33 | 84.62 | 37.50 |
| F3 | F1 | 80.00 | 92.31 | 75.00 |

Similarly, Table 3, has WRR obtained for the same dataset but by using Bing Speech API.

TABLE III. WRR for Bing Speech:

| Speaker 1 | Speaker 2 | Script1 | Script2 | Script3 |
|---|---|---|---|---|
| M1 | M2 | 100.00 | 92.31 | 87.50 |
| M2 | M3 | 93.33 | 84.62 | 87.50 |
| M3 | M1 | 86.67 | 92.31 | 93.75 |
| M2 | M4 | 86.67 | 84.62 | 81.25 |
| F1 | F2 | 80.00 | 84.62 | 93.75 |
| F2 | F3 | 93.33 | 92.31 | 87.50 |
| F3 | F1 | 86.67 | 76.92 | 93.75 |

The average of the WRR obtained from the previous table is given in Table 4.

TABLE IV. Average:

| Speech Engine | Average for Script1 | Average for Script2 | Average for Script3 | Average |
|---|---|---|---|---|
| Sphinx4 | 38.10 | 30.77 | 36.61 | 35.16 |
| Google Speech API | 86.67 | 85.71 | 66.07 | 79.48 |
| Bing Speech API | 89.52 | 86.81 | 89.29 | 88.54 |

*B. Results for Speaker Discrimination System:*

The accuracy of the speaker identification with respective to the number of features is illustrated in Figure 5. The number of features are varied from 1 to 26. Dynamic Time Wrapping is used as the feature mapping technique in our research. Various distance commutation methods such as Euclidean, Canberra are with DTW and compared. The accuracy vs number of features graph is shown in Graph3. It is noted that the system is highly accurate when we used 12 – 14 features, hence we took 13 features to process in the system

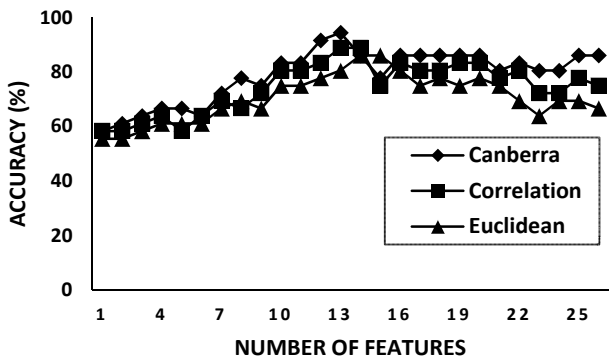

.

Fig. 6. Accuracy vs Number of Features

*C. Results for Sentiment Analysis System:*

Table 5, shows the accuracy of a different algorithms used for sentiment analysis such as Naive Bayes, Linear SVM, VADER.

TABLE V. Accuracy Sentiment

| Method | Twitter Dataset | Movie Review |
|---|---|---|
| Naive Bayes | 84 | 72.8 |
| Linear SVM | 88 | 86.4 |
| VADER | 95.2 | 96 |

VI. CONCLUSION AND FUTURE WORK

This work presents a generalized model that takes an audio which contains a conversation between two people as input and studies the content and speakers' identity by automatically converting the audio into text and by performing speaker recognition. In this research, we have proposed a simple system to do the above-mentioned task. The system works well with the artificially generated dataset, we are working on collecting a larger dataset and increasing the scalability of the system. Though the system is accurate in comprehending the sentiment of the speakers in conversational dialogue, it suffers some flaws, right now the system can handle a conversation between two speakers and in the conversation only one speaker should talk at a given time, it cannot understand if two people talk simultaneously. Our future work would address these issues and improve the accuracy and scalability of the system.